\documentclass{article}
\usepackage{spconf,amsmath,graphicx}
\usepackage{amssymb}
\usepackage{multirow}
\usepackage[normalem]{ulem}
\useunder{\uline}{\ul}{}
\usepackage{subfigure}
\usepackage{color}
\usepackage{lineno,hyperref}
\usepackage{booktabs}
\usepackage{lscape}
\usepackage{array}
\usepackage{float}
\usepackage{balance}
\usepackage{caption}
\title{METACOGNITION-ENHANCED FEW-SHOT PROMPTING WITH POSITIVE REINFORCEMENT}

\name{Yu Ji$^{1,2}$, Wen Wu$^{2,3,}$\sthanks{Correspongding author: wwu@cc.ecnu.edu.cn}, Yi Hu$^{3}$, Hong Zheng$^{4,}$\sthanks{Correspongding author: zhhmm2@163.com}, Liang He$^{1,2}$}
\address{$^{1}$Institute of AI for Education, East China Normal University, Shanghai, China \\
$^{2}$School of Computer Science and Technology, East China Normal University, Shanghai, China \\
$^{3}$Shanghai Key Laboratory of Mental Health and Psychological Crisis Intervention, \\ School of Psychology and Cognitive Science, East China Normal University, Shanghai, China \\
$^{4}$Shanghai Changning Mental Health Center, Shanghai, China}

\begin{document}
\maketitle

\begin{abstract}
Few-shot prompting elicits the remarkable abilities of large language models by equipping them with a few demonstration examples in the input. However, the traditional method of providing large language models with all demonstration input-output pairs at once may not effectively guide large language models to learn the specific input-output mapping relationship. In this paper, inspired by the regulatory and supportive role of metacognition in students' learning, we propose a novel metacognition-enhanced few-shot prompting, which guides large language models to reflect on their thought processes to comprehensively learn the given demonstration examples. Furthermore, considering that positive reinforcement can improve students' learning motivation, we introduce positive reinforcement into our metacognition-enhanced few-shot prompting to promote the few-shot learning of large language models by providing response-based positive feedback. The experimental results on two real-world datasets show that our metacognition-enhanced few-shot prompting with positive reinforcement surpasses traditional few-shot prompting in classification accuracy and macro F1.
\end{abstract}


\section{Introduction}\label{sec-introduction}

\begin{figure}
	\centering
	\includegraphics[width=3.4in]{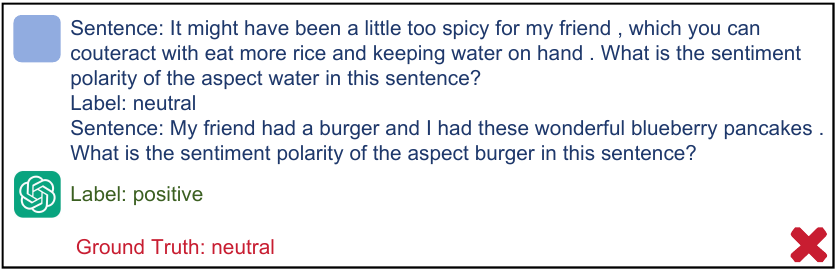}
	\caption{Example of the weakness of few-shot prompting.}\label{figure-pre-example}
\end{figure}

Recently, Large Language Models (LLMs) (e.g., ChatGPT, GPT-4, LLaMA 2, and Claude 2) have exhibited impressive capabilities in various downstream tasks (e.g., sentiment analysis \cite{wang2023chatgpt} and personality prediction \cite{ji2023chatgpt}) with the assistance of different prompting strategies (e.g, few-shot prompting \cite{min2022rethinking}). As one of the commonly employed prompting strategies, few-shot prompting provides a few demonstration examples of desired input-output pairs in the input of LLMs, which enables LLMs to quickly learn the specific input-output mapping relationship corresponding to the downstream task \cite{brown2020language}. However, this passive learning of the given demonstration examples (similar to spoon-feeding in education \cite{dehler2014against}) makes LLMs lack the autonomous reflection of their thought processes, which may limit their cognitive development and consequently affect their performance in downstream tasks. Take Apsect-Based Sentiment Classification (ABSC) task \cite{cheng2023tell} as an example, we select two similar samples from 14-Restaurant dataset, with one serving as a demonstration example and the other as a test sample. As shown in Fig.~\ref{figure-pre-example}, even if few-shot prompting provides the demonstration input-output pair for ChatGPT to learn, ChatGPT still makes wrong prediction for the test sample which is similar to the given demonstration example. This phenomenon indicates that only providing demonstration input-output pairs directly to LLMs may not necessarily help LLMs learn the specific mapping relationship behind the given demonstration examples.

In this paper, we are motivated to propose a novel MetaCognition-enhanced Few-Shot (MCeFS) prompting to improve the performance of traditional few-shot prompting, where the idea is inspired by the regulatory and supportive role of learners' metacognition in their learning \cite{stanton2021fostering}. Concretely, metacognition refers to thinking about thinking, which is the individual's ability to plan, monitor, evaluate, and reflect on her/his own learning \cite{guo2022using}. Recently, some educational psychologists have attempted to improve students' learning performance in classroom by enhancing their metacognition \cite{darmawan2020simas}. Similarly, we design MCeFS prompting to enhance the metacognition of LLMs to better accomplish downstream tasks. Concretely, MCeFS prompting requires LLMs to analyze the given demonstration examples one by one and reflect on their thought processes about the analysis of demonstration examples, thus enabling LLMs to better understand the specific mapping relationship behind the given demonstration examples. Furthermore, considering that teachers normally use positive reinforcement (i.e., providing rewards to increase the frequency of good learning behaviors) to enhance students' learning motivation \cite{sumiati2019building}, we introduce positive reinforcement into our MCeFS prompting. To be specific, we offer appropriate positive feedback to LLMs based on their analysis results of the given demonstration examples, which could guide LLMs to develop their thinking towards accurately completing specific downstream task.

The main contributions of our work are as follows:

(1) We have proposed a novel MCeFS prompting to better elicit the abilities of LLMs with a few demonstration examples. Compared with traditional few-shot prompting, our MCeFS prompting could guide LLMs to learn the given demonstration examples more comprehensively.

(2) We have introduced positive reinforcement into the few-shot learning of LLMs. By providing positive feedback corresponding to the responses of LLMs, LLMs are promoted to accomplish downstream tasks more precisely.

(3) We have conducted experiments on two real-world datasets to verify the performance of our MCeFS prompting with positive reinforcement. The experimental results illustrate that our MCeFS prompting with positive reinforcement outperforms traditional few-shot prompting in terms of classification accuracy and macro F1.

\section{Related Work}\label{sec-related-work}

{\bf Few-shot Prompting.} With the rapid development of LLMs, more and more researchers attempted to develop and optimize different prompting strategies to effectively employ LLMs in various downstream tasks (e.g., sentiment analysis \cite{wang2023chatgpt} and personality prediction \cite{ji2023chatgpt}) \cite{zhou2022large}. Among the various prompting strategies, few-shot prompting is one of the simple and effective prompt strategies, which allows LLMs to capture the specific mapping relationship corresponding to downstream task by providing a few demonstration examples in the input of LLMs \cite{min2022rethinking,brown2020language}. Therefore, some researchers attempted to optimize few-shot prompting for better eliciting the abilities of LLMs \cite{lu2022fantastically,liu2022makes,ma2023fairness}. For example, Lu et al. \cite{lu2022fantastically} adopted the generative nature of language models to construct demonstration examples and utilized entropy-based statistics to sort the demonstration examples. Liu et al. \cite{liu2022makes} proposed to retrieve demonstration examples that are semantically-similar to a test sample to construct its corresponding prompting. Ma et al. \cite{ma2023fairness} introduced a metric to evaluate the predictive bias of a fixed prompting against labels and designed greedy search based strategy to select prompting. However, most of them normally provide all demonstration input-output pairs for LLMs to learn. In fact, in this passive learning approach, LLMs may only learn the specific output format rather than the deeper input-output mapping relationship.

{\bf Metacognition.} Metacognition refers to the ability to think about and reflect on one's own cognitive process \cite{guo2022using}. Previous studies have shown that metacognition plays a critical role in successful learning \cite{mahdavi2014overview}. Therefore, an increasing number of researchers tried to enhance individuals' metacognition to assist them in learning more effectively. For example, Meteier et al. \cite{meteier2023enhancing} improved the learning of nursing students by enhancing their metacognition with eye-tracking glasses. Ward et al. \cite{ward2021orgbox} designed a search-assistance tool to encourage users to engage in metacognitive activities during their search processes (e.g., reflect on the information they found). However, most researchers focus on improving individuals' metacognition while overlooking the impact of metacognition on the learning of LLMs.

{\bf Positive Reinforcement.} Positive reinforcement involves the usage of desirable or pleasant stimuli after the performance of certain behaviors to increase the likelihood that the behaviors will occur \cite{hardy2020using}. The field of education is one of the common fields where positive reinforcement is widely employed. Concretely, teachers often use positive reinforcement (e.g., praise or other verbal reinforcement, tangible rewards, and token rewards \cite{adamson2015understanding,fitriati2020teachers}) to help boost students' learning motivation, thus guiding them to learn more effectively \cite{sumiati2019building}. 

Hence, we would like to enhance the metacognition of LLMs to guide them to better learn the given demonstration examples. Besides, we are interested in introducing positive reinforcement into the few-shot learning of LLMs to promote them to better complete downstream tasks.  

\section{Methodology}\label{sec-methodology}
\begin{figure}
	\centering
	\includegraphics[width=3.4in]{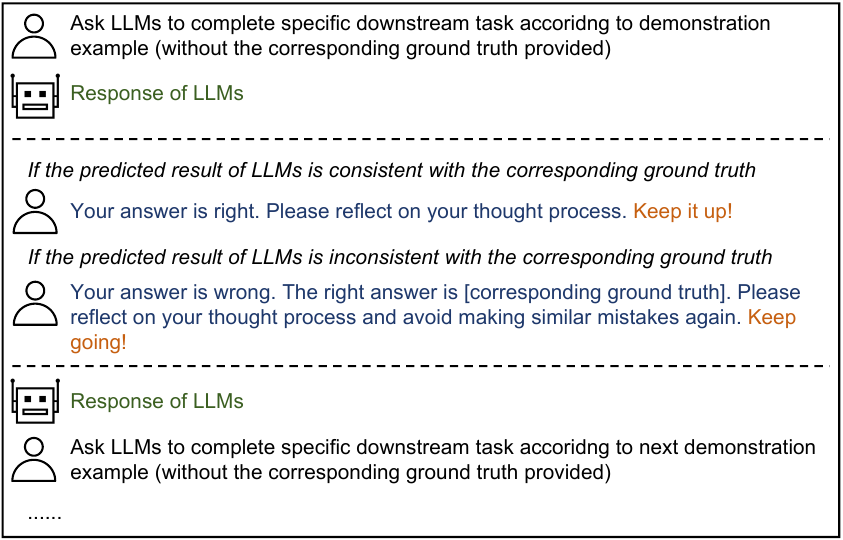}
	\caption{The learning process of our MCeFS prompting with positive reinforcement. The blue and orange parts are the core of MCeFS prompting and positive reinforcement respectively.}\label{figure-framework}
\end{figure}

In this section, we will introduce the details of our proposed MCeFS prompting with positive reinforcement. Fig.~\ref{figure-framework} shows the learning process of our MCeFS prompting with positive reinforcement. To be specific, for each demonstration example, we no longer provide the corresponding input-output pair in the input of LLMs. Instead, we ask LLMs to complete the specific downstream task according to the given demonstration example, while the corresponding ground truth is not provided. We then assess the prediction result of LLMs. If the prediction result of LLMs is consistent with the corresponding ground truth, we will praise them (i.e., one type of positive reinforcement) and ask them to reflect on their thought processes. Otherwise, we will require them to reflect on their thought processes as well and urge them to avoid making comparable errors again. Note that, for the employment of positive reinforcement, we request LLMs to provide several common praises (e.g., you're really good) and use them to simulate the learning motivation of LLMs. Finally, we select the best one according to the results.

\section{Experiments}\label{sec-experiments}
In this section, we conduct experiments on two real-world datasets to validate the effect of MCeFS prompting and positive reinforcement\footnote{Code is available at https://github.com/jiyu0201/MCeFSL.}. Specifically, we aim to answer the following Research Questions (RQs):

{\bf RQ1}: Whether our MCeFS prompting could guide LLMs to better learn the given demonstration examples than traditional few-shot prompting? (see Section~\ref{sec-rq1})

{\bf RQ2}: Whether positive reinforcement is effective for promoting the few-shot learning of LLMs? (see Section~\ref{sec-rq2})

\subsection{Datasets}\label{sec-datasets}

For a fair comparison, we choose the same downstream task (i.e., ABSC task) as Wang et al. \cite{wang2023chatgpt}. Similarly, we conduct experiments on SemEval-2014 Datasets\footnote{https://alt.qcri.org/semeval2014/task4/} (i.e., 14-Laptop and 14-Restaurant). The statistics of the two datasets are shown in Table~\ref{table-dataset}. Besides, we use \textit{Accuracy} and \textit{Macro F1} as evaluation metrics \cite{grandini2020metrics}.

\subsection{Implementation Details}\label{sec-implementation-details}

For the usage of LLMs, we adopt the representative version of ChatGPT (i.e., gpt-3.5-turbo). In addition, we set the temperature to 0 to produce more deterministic and focused responses. To minimize the variance resulting from the sampling of demonstration examples, we adopt three random seeds (i.e., 13, 42, 550) for sampling to conduct experiments and report the average performance. Considering the limitations of ChatGPT on the input length, the shot number is selected from [1,\;3,\;9]. Moreover, we utilize the zer-shot prompting\footnote{Zer-shot prompting: \textit{Sentence: \{sentence\} What is the sentiment polarity of the aspect \{aspect\} in this sentence?}} designed by Wang et al. \cite{wang2023chatgpt} to request LLMs to complete ABSC task. 

\begin{table}
\small
\centering
\caption{Statistics of 14-Laptop and 14-Restaurant datasets}\label{table-dataset}
\begin{tabular}{lcc}
\hline
Dataset       & \#Train & \#Test \\ \hline
14-Laptop     & 2,282   & 632    \\ 
14-Restaurant & 3,608   & 1,119  \\ \hline
\end{tabular}
\end{table}

\begin{table}
\small
\centering
\caption{The experiment results about the effectiveness of MCeFS prompting. The boldface indicates the best model results of the dataset, and the underline indicates the second best model result of the dataset}\label{table-rq1}
\begin{tabular}{lcccc}
\hline
\multicolumn{1}{l}{\multirow{2}{*}{Model}} & \multicolumn{2}{c}{14-Laptop}                                             & \multicolumn{2}{c}{14-Restaurant}                    \\ \cline{2-5} 
\multicolumn{1}{l}{}                       & \multicolumn{1}{c}{Accuracy}        & \multicolumn{1}{c}{Macro F1}       & \multicolumn{1}{c}{Accuracy}        & Macro F1       \\ \hline
\multicolumn{5}{l}{\textit{Fully-supervised models}}                                                                                                                                     \\ \hline
\multicolumn{1}{l}{SOTA}                   & \multicolumn{1}{c}{\textbf{83.7\%}} & \multicolumn{1}{c}{\textbf{0.801}} & \multicolumn{1}{c}{\textbf{89.5\%}} & \textbf{0.849} \\ \hline
\multicolumn{5}{l}{\textit{ChatGPT with few-shot prompting}}                                                                                                                 \\ \hline
\multicolumn{1}{l}{Few-Shot (1)}           & \multicolumn{1}{c}{74.5\%}          & \multicolumn{1}{c}{0.622}          & \multicolumn{1}{c}{82.3\%}          & 0.640          \\ 
\multicolumn{1}{l}{Few-Shot (3)}           & \multicolumn{1}{c}{75.5\%}          & \multicolumn{1}{c}{0.641}          & \multicolumn{1}{c}{83.2\%}          & 0.671         \\ 
\multicolumn{1}{l}{Few-Shot (9)}           & \multicolumn{1}{c}{74.4\%}          & \multicolumn{1}{c}{0.613}          & \multicolumn{1}{c}{83.2\%}          & 0.667          \\ \hline
\multicolumn{5}{l}{\textit{ChatGPT with our MCeFS prompting}}                                                                                                                                    \\ \hline
\multicolumn{1}{l}{MCeFS (1)}             & \multicolumn{1}{c}{79.5\%}        & \multicolumn{1}{c}{{\ul 0.746}}  & \multicolumn{1}{c}{84.7\%}          & 0.743          \\ 
\multicolumn{1}{l}{MCeFS (3)}             & \multicolumn{1}{c}{{\ul 80.3\%}}    & \multicolumn{1}{c}{0.745}         & \multicolumn{1}{c}{84.2\%}          & {\ul 0.775}  \\ 
\multicolumn{1}{l}{MCeFS (9)}             & \multicolumn{1}{c}{79.0\%}         & \multicolumn{1}{c}{0.722}        & \multicolumn{1}{c}{{\ul 86.0\%}}    & 0.773         \\ \hline
\end{tabular}
\end{table}

\subsection{Effectiveness of MCeFS Prompting ({\bf RQ1})}\label{sec-rq1}

The related experimental results are shown in Table~\ref{table-rq1}, where the performance of the State-Of-The-Art (SOTA) models on 14-Laptop and 14-Restaurant datasets are quoted from \cite{li2022modeling} and \cite{zhang2022towards}. In addition, Few-Shot (k) and MCeFS (k) represent ChatGPT with traditional few-shot prompting and our MCeFS prompting respectively, while $k$ stands for the shot number. It can be seen from Table~\ref{table-rq1} that compared with few-shot prompting, our MCeFS prompting better elicits the sentiment analysis ability of ChatGPT and largely reduces the performance gap between ChatGPT and SOTA. Concretely, under the same shot number, our MCeFS prompting has better performance than traditional few-shot prompting w.r.t. classification accuracy and macro F1 on the two datasets. For example, relative to Few-Shot (3), MCeFS (3) increases its macro F1 from 0.671 to 0.775 on 14-Restaurant dataset. Furthermore, even if our MCeFS prompting uses fewer demonstration examples than traditional few-shot prompting, the performance of our MCeFS prompting still surpasses that of traditional few-shot prompting. For instance, the classification accuracy of MCeFS (3) has been improved from 74.4\% to 80.3\% on 14-Laptop when compared to Few-Shot (9). As for the reason, traditional few-shot prompting allows LLMs to learn the given demonstration input-output pairs by the way of passively receiving information, which may limit the autonomous reflection and thinking development of LLMs. In contrast, by guiding LLMs to reflect on their own thought processes regarding the completion of the downstream task, our MCeFS prompting enhances the metacognition of LLMs, which enables LLMs to understand the mapping relationship corresponding to the downstream task more precisely. Hence, LLMs could possess a more targeted ability to solve the downstream task with a few demonstration examples.

\begin{figure}
	\centering
	\subfigure[14-Laptop \& Accuracy]{\includegraphics[width=1.65in]{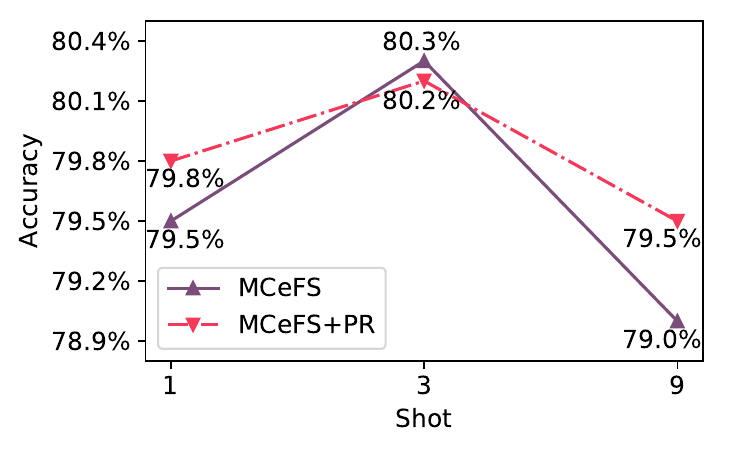}}
    \subfigure[14-Laptop \& Macro F1]{\includegraphics[width=1.65in]{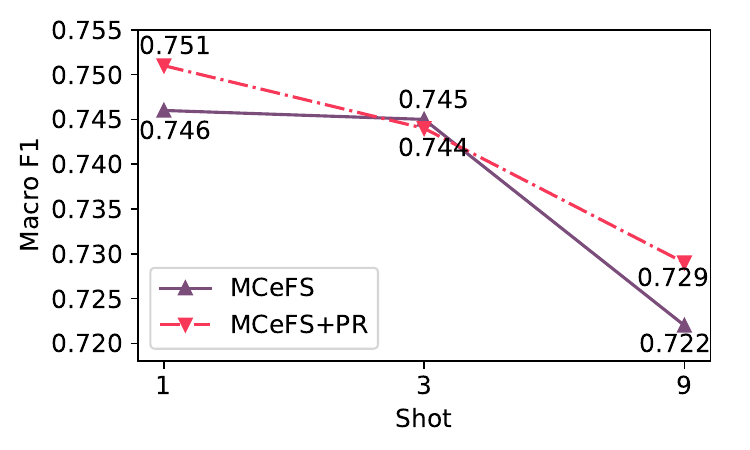}}

\subfigure[14-Restaurant \& Accuracy]{\includegraphics[width=1.65in]{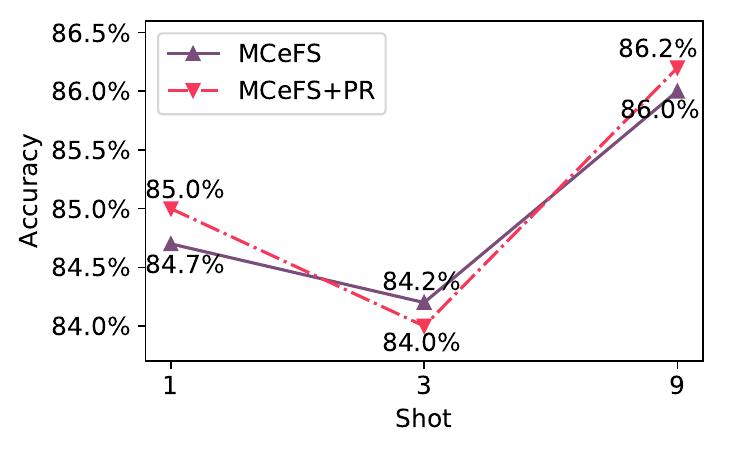}}
\subfigure[14-Restaurant \& Macro F1]{\includegraphics[width=1.65in]{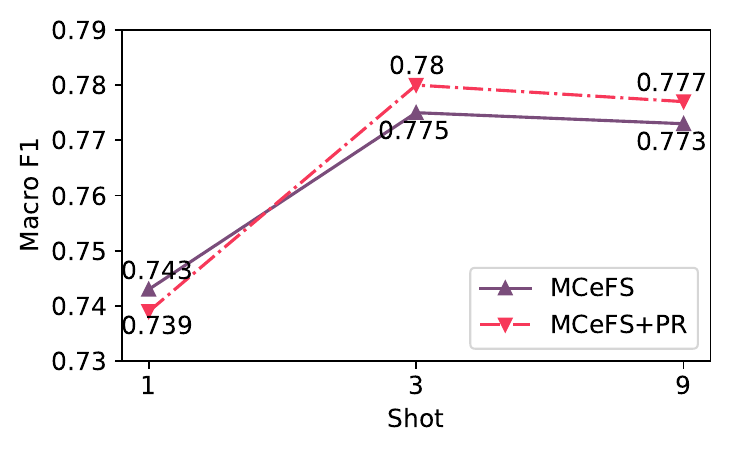}}
	\caption{The experiment results about the effectiveness of positive reinforcement.}\label{fig-rq2}
\end{figure}

\subsection{Effectiveness of Positive Reinforcement ({\bf RQ2})}\label{sec-rq2}

The related experimental results are shown in Fig.~\ref{fig-rq2}, where MCeFS+PR denotes MCeFS prompting with Positive Reinforcement. Concretely, we observe that under the same shot number, the performance of our MCeFS prompting has been further enhanced after incorporating positive reinforcement in most cases. For instance, compared with MCeFS (1), MCeFS+PR (1) increases its macro F1 from 0.746 to 0.751 on 14-Laptop dataset. The possible reason is that positive reinforcement offers response-based positive feedback (i.e., praise) to simulate the learning motivation of LLMs, enabling them to learn in the direction of accurately analyzing given demonstration examples, thereby gaining a deeper understanding of the specific mapping relationship behind the given demonstration examples and ultimately completing the downstream task precisely. This discovery offers new insights into guiding LLMs learning from a psychological perspective.

\subsection{Case Study}\label{sec-case-study}

\begin{figure}
	\centering
	\includegraphics[width=3.4in]{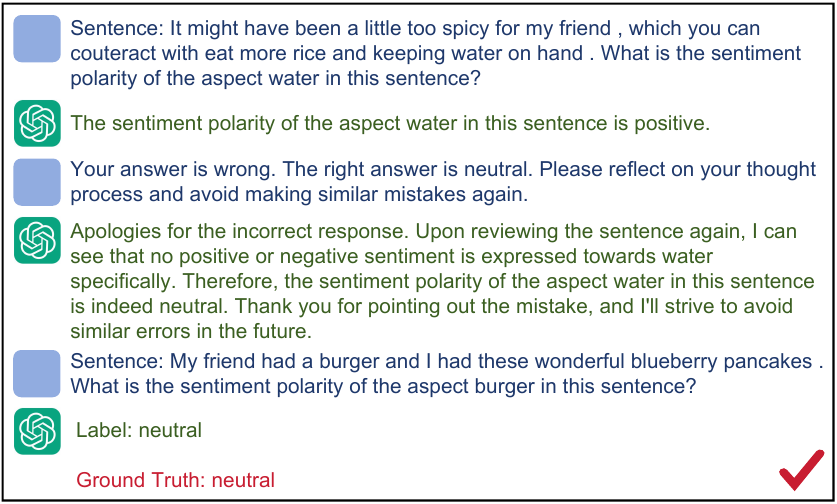}
	\caption{Example of our MCeFS prompting.}\label{fig-pre-example-mcefs}
\end{figure}

To illustrate the effectiveness of our MCeFS prompting, we take the samples used in Fig.~\ref{figure-pre-example} to display the learning process of ChatGPT with our MCeFS prompting. It can be seen from Fig.~\ref{fig-pre-example-mcefs}, ChatGPT does not analyze the given demonstration example correctly. Subsequently, with the guidance of our MCeFS prompting, ChatGPT reflects on its thought process and recognizes its mistake. Finally, when encountering the test sample that is similar to the demonstration example, ChatGPT successfully makes correct prediction for the test sample. This case study shows that our MCeFS prompting could enhance the performance of LLMs on downstream tasks by guiding LLMs to autonomous reflection. In addition, the reflective process presented by LLMs also benefits researchers in uncovering cognitive errors in LLMs.

\section{Conclusions and future work}\label{sec-conclusions}

To address the problem of insufficient learning from demonstration examples in traditional few-shot prompting, we propose a novel metacognition-enhanced few-shot prompting to guide LLMs in better learning the given demonstration examples. Furthermore, we introduce positive reinforcement into our MCeFS prompting to promote LLMs learning towards accurate completion of downstream tasks. The experimental results on two real-world datasets show that our MCeFS prompting with positive reinforcement could better elicit the abilities of LLMs than traditional few-shot prompting. As for future work, we are interested in designing a Human-In-The-Loop (HTIL) learning framework and introducing expert feedback corresponding to the reflective process of LLMs to further develop their thinking.

\section*{Acknowledgment}
This work is funded by National Natural Science Foundation of China (under project No. 62377013), Science and Technology Commission of Shanghai Municipality, China (under project No. 21511100302), Natural Science Foundation of Shanghai (under project No. 22ZR1419000), the Research Project of Changning District Science and Technology Committee (under project No. CNKW2022Y37), and the Medical Master's and Doctoral Innovation Talent Base Project of Changning District (under project No.RCJD2022S07). 

\bibliographystyle{IEEEbib}
\bibliography{refs}

\end{document}